\documentclass{article}
\usepackage{main}

\usepackage{times}
\usepackage{soul}
\usepackage{url}
\usepackage[hidelinks]{hyperref}
\usepackage[utf8]{inputenc}
\usepackage[small]{caption}
\usepackage{graphicx}
\usepackage{amsmath}
\usepackage{amsthm}
\usepackage{booktabs}
\usepackage{algorithm}
\usepackage{algorithmic}
\usepackage[switch]{lineno}

\title{Unreliable Partial Label Learning with Recursive Separation}

\author{
Yu Shi
\and
Ning Xu\footnote{Corresponding authors.} \and
Hua Yuan \And
Xin Geng$^{*}$
\affiliations
School of Computer Science and Engineering, Southeast University, Nanjing 211189, China
\emails
\{seushiyu, xning, yuanhua, xgeng\}@seu.edu.cn
}

\usepackage{bbding}
\usepackage{multirow}
\usepackage{dsfont}

\begin{document}

\maketitle

\begin{abstract}
Partial label learning (PLL) is a typical weakly supervised learning problem in which each instance is associated with a candidate label set, and among which only one is true.
However, the assumption that the ground-truth label is always among the candidate label set would be unrealistic, as the reliability of the candidate label sets in real-world applications cannot be guaranteed by annotators.
Therefore, a generalized PLL named Unreliable Partial Label Learning (UPLL) is proposed, in which the true label may not be in the candidate label set. 
Due to the challenges posed by unreliable labeling, previous PLL methods will experience a marked decline in performance when applied to UPLL.
To address the issue, we propose a two-stage framework named Unreliable Partial Label Learning with Recursive Separation (UPLLRS).
In the first stage, the self-adaptive recursive separation strategy 
is proposed to separate the training set into a reliable subset and an unreliable subset.
In the second stage, a disambiguation strategy is employed to progressively identify the ground-truth labels in the reliable subset.
Simultaneously, semi-supervised learning methods are adopted to extract valuable information from the unreliable subset.
Our method demonstrates state-of-the-art performance as evidenced by experimental results, particularly in situations of high unreliability.
Code and supplementary materials are available at \url{https://github.com/dhiyu/UPLLRS}.
\end{abstract}

\section{Introduction}

\label{introduction}
Partial label learning (PLL) is a typical weakly supervised learning problem where the candidate label set is given for each instance but among which only one is true. 
Compared with the ordinary supervised learning problem where each instance is associated with only one ground-truth label, 
partial label learning induces predictive model from ambiguous labels, hence considerably reduces the cost of data annotations. 
Nowadays, PLL has been extensively employed in the field of web mining \cite{luo2010learning}, multimedia content analysis \cite{zeng2013learning}, 
automatic image annotations \cite{chen2018learning}, ecoinformatics \cite{liu2012conditional,tang2017confidence}, etc.

A variety of methods have been proposed for addressing the PLL problem.
The most common strategy to learn from partial labels is disambiguation, where
Identification-Based Strategy (IBS) and Average-Based Strategy (ABS) are two main disambiguation strategies.
For IBS, iterative optimization is employed to predict true label treated as latent variable.
While ABS treats all labels in the candidate label set in an equal manner where probabilities of modeling outputs are averaged to get the final prediction.
It memorizes all candidate labels, since it avoids identifying the latent ground truth label.
Recently, deep neural network based IBS method have achieved promising performance on PLL. 
PiCO \cite{wang2022pico} achieves a significant improvement in performance by adopting contrastive learning strategy in PLL, which is able to learn high quality representation.
CR-DPLL \cite{wu2022revisiting} is a novel consistency regularization method which achieved state-of-the-art performance on PLL, almost nearing supervised learning.
Even so, whether IBS or ABS, both assumed that the true label is present within the candidate label set. 

However, the assumption that the true labels are consistently present within the candidate label sets would be unrealistic.
In existing PLL setting, the annotation for each instance is the partial labels (i.e. candidate label set) rather than the true label directly, thus significantly reduces the difficulty and cost.
Against this backdrop, it poses a challenge for annotators to ensure that the true labels are present within the candidate label sets.
Therefore, Unreliable Partial Label Learning (UPLL) \cite{lv2021robustness} is proposed in response, which is a more general problem than existing PLL. 
In UPLL, it is acknowledged that the true label may not be present within the candidate label set for each instance.
Since that, it significantly reduces the difficulty and cost associated with data annotation.
Above all, UPLL addresses the issue of labeling instances that are difficult to distinguish.
Hence, UPLL could be deemed a more prevalent and valuable problem.

Despite existing PLL methods achieved promising performance, suffering from unreliable partial labeling, current PLL methods encounter numerous challenges when applied to UPLL.
It will exhibit a significant decline in performance on UPLL datasets, particularly for high unreliable rates.
RABS \cite{lv2021robustness} has demonstrated that bounded loss functions have the ability to fit the ground-truth label against the interference of unreliability and other candidates.
However, it would fail on high unreliable levels or high partial levels.
This urges us to design an efficient method to manipulate high unreliability problem.

Motivated by this consideration, a framework named Unreliable Partial Label Learning with Recursive Separation (UPLLRS) is devised to issue this puzzle.
In this paper, a novel separation method named Recursive Separation (RS) is proposed for known unreliable rate scenes to separate unreliable samples and reliable samples. 
However, it is limited in the real world because the real unreliable rate is difficult to know. 
In order to tackle this problem, more generally, we design a self-adaptive strategy for RS algorithm which could fit unknown unreliable rate. 
Pilot experiments have demonstrated the effectiveness of self-adaptable RS algorithm.
After that, we combine a label disambiguation strategy with semi-supervised learning techniques in the second stage of UPLLRS.
Experiments show that our method achieve state-of-the-art results on the UPLL datasets.
Our contributions can be summarized as follows:
\begin{itemize}
    \item A self-adaptive recursive separation algorithm is proposed for effectively separating raw dataset into a reliable subset and an unreliable subset. 
    \item A two-stage framework is proposed for inducing the predictive model, based on the self-adaptive RS strategy.
    Upon obtaining both the reliable subset and unreliable subset, the disambiguation strategy utilizes the reliable subset for learning while incorporating information from the unreliable subset by the semi-supervised technique.
    \item The UPLLRS framework is versatile, capable of handling both image and non-image datasets.
    Utilizing data augmentation techniques on image datasets, the performance will be further enhanced.
\end{itemize}

The rest of this paper is organized as follows. 
First, we briefly review related works on partial label learning.
Second, the details of the proposed UPLLRS are introduced.
Third, we present the results of the comparative experiments, followed by the final conclusion.

\section{Related Work}

Partial label learning deals with the problem that the true label of each instance resides in the candidate label set. 
Many algorithms have been proposed to tackle this problem, with existing PLL methods broadly classified into classical and deep learning approaches.

In classical PLL, label disambiguation is based on averaging or identification. In averaging-based methods, the candidate label set and non-candidate label set are treated the same\cite{he2006ida,cour2011jmlr,zhang2015ijcai}. 
For example, \cite{cour2011jmlr} discriminated candidate labels and non-candidate labels with a convex loss.
But identification-based methods progressively refine labels in the candidate set during the model training\cite{chen2013cvpr,yu2016acml}. 
\cite{yu2016acml} optimized the constraint on the maximum margin between maximum modeling output of candidate labels and that of other labels.

However, the model output of averaging-based methods often overwhelms the true label, resulting in low accuracy. As a result, many identification-based algorithms have been devised in recent years\cite{feng2019aaai,gong2018trans,lyu2021tkde,tang2017confidence,xu2019plle}.
Nevertheless, these classical methods often have a bottleneck due to the restriction of the linear model.

Given the success of deep neural network-based methods in classification tasks, a proliferation of PLL approaches incorporating deep neural networks have emerged.
\cite{yao2020a} designed two regularization techniques in the training with ResNet representing the first exploration of deep PLL. 
\cite{yao2020b}, referring to the idea of co-training, trained two networks to interact with each other for label disambiguation. 
Concurrently, a progressive method proposed by \cite{lv2020progressive} progressively identified true label adopting the memorization effect of deep network. 
\cite{feng2020provably} formalized the partial label generation process and proposed two provably consistent algorithms, risk consistent (RC) classifier and classification consistent (CC) classifier. 
Then, a leveraged weighted loss, which balances the contributions of candidate labels and non-candidate labels, was proposed by \cite{wen2021leveraged}. 
With the development of contrastive learning, \cite{wang2022pico} applied contrastive learning to PLL for effective feature representation. 
Recently, \cite{wu2022revisiting} designed a consistency regularization framework in deep PLL, which gives very small performance drop compared with fully supervised learning.

However, it is common for false positive labels to be inadvertently chosen from label set, rather than being selected randomly.
More specifically, it is acknowledged that each instance may not possess a uniform prior label distribution, but rather a latent label distribution which encompasses vital labeling information.
Thus, \cite{xn2021valen} proposed an instance-dependent approach named VALEN, which aims to recover the latent label distribution via label enhancement \cite{xu2022variational,xu2019label}, leveraging it to further improve performance in real-world settings.
VALEN first generates a label distribution through label enhancement, then utilizes variational inference to approximate that distribution.

In practice, despite the demonstrated empirical success of the aforementioned algorithms in the PLL task, their effectiveness is limited when the ground-truth label may not be present within the candidate label set.
Therefore, the unreliable partial label learning is proposed in \cite{lv2021robustness}, which is more general in comparison to current PLL.
Furthermore, \cite{lv2021robustness} has proved that bounded loss, such as the Mean Absolute Error (MAE) loss and the Generalized Cross Entropy (GCE) loss, is robust against unreliability. However, the performance of RABS remains limited in the presence of high levels of unreliability.

\section{Preliminaries}

Let $\mathcal{X}$ and $\mathcal{Y}$ be feature space and label space respectively, 
and 
$p(\boldsymbol{x}, y)$ be the distribution on $\mathcal{X} \times \mathcal{Y}$.
Moreover,
$D=\{(\boldsymbol{x}_i, {y}_i)\}_{i=1}^{n}$ is the training set in which $\boldsymbol{x}_i$ is $i$-th instance and $y_i$ is the corresponding ground-truth label, 
and $V=\{(\boldsymbol{x}_i, y_i)\}_{i=1}^{k}$ is the validation set including $k$ pairs of instance $\boldsymbol{x}_i$ with ground-truth label $y_i$.

In PLL problem, the distribution $p(\boldsymbol{x}, y)$ is corrupted to $p(\boldsymbol{x}, s)$ in which $s$ is candidate label set satisfied $p (y_i \in s_i)=1$, $\forall y_i \in \mathcal{Y}$,
and the training set is corrupted to $\bar{D}=\{(\boldsymbol{x}_i, s_i)\}_{i=1}^{n}$.
The goal of PLL task is to induce a classifier from ambiguous dataset $\bar{D}$.

However, in UPLL, $\mu$ called unreliable rate is the probability of true label $y_i$ not in the candidate label set $s_{i}$, it can be expressed formally as:
\begin{equation}
	p(y_i \in s_i)=1-\mu.
\end{equation}
Such that the candidate label set $s_{i}$ is corrupted to $\tilde{s}$ that is the unreliable candidate label set. Then, the UPLL training set can be denoted as $\tilde{D}=\{(\boldsymbol{x}_i, \tilde{s}_i)\}_{i=1}^{n}$.

\section{Proposed Method}

In this section, we firstly propose the self-adaptive Recursive Separation (RS) algorithm which aims to differentiate reliable samples and unreliable samples effectively.
After that, numerous pilot experiments were conducted which demonstrate that self-adaptive RS algorithm is effective.
Finally, a framework entitled Unreliable Partial Label Learning with Recursive Separation (UPLLRS) is proposed. 
There are two key stages in this framework. 
At the beginning, the self-adaptive Recursive Separation (RS) effectively split training dataset into a reliable subset and an unreliable subset. 
Subsequently, in order to induce a predictive model, a disambiguation strategy is employed to progressively identify ground-truth labels while utilizing semi-supervised learning techniques in combination.
\begin{algorithm}[tb]
    \caption{Self-adaptive RS Algorithm}
    \label{alg:RS}
    \textbf{Input}: Separation network $f(\cdot; \theta)$ with trainable parameters $\theta$;
    Unreliable partial label training set $\tilde{D}=\{(\boldsymbol{x}_i, \tilde{s}_i)\}_{i=1}^{n}$ and validation set $V=\{(\boldsymbol{x}_i, y_i)\}_{i=1}^{k}$;
    Small epochs $\beta$ for each separation step;
    Separation rate $\gamma$;
    RS patience $\varphi$ and max separation step $\lambda$. \\ 
    \textbf{Output}: Reliable subset $\tilde{D}_{R}^{\lambda}=\{(\boldsymbol{x}_i, \tilde{s}_i)\}_{i=1}^{m}$ and unreliable subset $\tilde{D}_{U}^{\lambda}=\{(\boldsymbol{x}_i)\}_{i=1}^{n-m}$.

    \begin{algorithmic}[1] 
    \STATE Let $\varphi_{\text{curr}} \leftarrow 0$ and $Acc_V \leftarrow 0$;
    \FOR{$i \leftarrow 1$ to $\lambda$}

        \STATE Randomly initialize $\theta_0^i$;
        \FOR{$j \leftarrow 1$ to $\beta$}
            \STATE Train $f(\cdot;\theta_{j-1}^i)$ using dataset $\tilde{D}_{R}^{i}$;
            \STATE Calculate loss $l$ according Eq. \ref{Eq.CCEloss};
            \STATE Update parameters from $\theta_{j-1}^{i}$ to $\theta_{j}^{i}$;
            \IF{$j = \beta$}
                \STATE Sort $l$ by value in descending order;
                \STATE Exclude top-$\gamma$ instances from $\tilde{D}_{R}^{i}$ and add excluded instances to $\tilde{D}_{U}^{i}$ without labels;
            \ENDIF

        \ENDFOR
        
        \STATE Evaluate $f(\cdot;\theta_{j}^i)$ on dataset $V$ and calculate accuracy $Acc_{\text{curr}}$;
        \IF {$Acc_{\text{curr}} < Acc_{V}$}
            \STATE $\varphi_{\text{curr}} \leftarrow \varphi_{\text{curr}}+1$;
            \IF {$\varphi_{\text{curr}} \ge \varphi$}
                \STATE break;
            \ENDIF
        \ELSE 
            \STATE $Acc_V \leftarrow Acc_{\text{curr}}$, $\varphi_{\text{curr}} \leftarrow 0$;
        \ENDIF

    \ENDFOR

    \STATE \textbf{return} Reliable subset $\tilde{D}_{R}^{\lambda}$ and unreliable subset $\tilde{D}_{U}^{\lambda}$.

    \end{algorithmic}
\end{algorithm}

\subsection{Recursive Separation}
\label{RS}

The memorization effect can be interpreted as the deep network firstly fit correct labels and then gradually fit wrong labels through the learning phase \cite{bai2021understanding}.
In recent years, small loss trick has been demonstrated to be an effective method for addressing label noise.
That inspires us to identify the reliability of samples and pay more attention to the reliable samples.
Furthermore, it is discovered that the top-10\% large loss samples contain more unreliable samples than any other parts after a few epochs of training, as Figure \ref{sample_num} shows.
This motivated us to progressively take unreliable partial samples away from the training set by iteratively excluding top-$\gamma$ large loss samples.
Based on this idea, we introduce a multi-class classifier $f(\cdot;\theta)$ with parameters $\theta$.
More specifically, the training phase of recursive separation task is optimizing the following classical multi-class classification objective function:
\begin{equation}
    \underset {\theta} { \operatorname {arg\,min} } \, \frac{1}{n}\sum_{i = 1}^{n}\mathcal{L}_{\text{RS}} (f(\boldsymbol{x}_i;\theta), s_i),
\end{equation}
where $\mathcal{L}_{\text{RS}}$ is the loss function for the Recursive Separation (RS) stage. According to \cite{liu2020early,bai2021understanding}, in the early-learning stage, the gradient direction of cross-entropy loss is close to the correct optimization direction. 
It inspires us to choose Categorical Cross Entropy (CCE) \cite{lv2021robustness} loss as $\mathcal{L}_{\text{RS}}$ under the UPLL setting, such that the objective function can be rewritten as:
\begin{equation}
    \label{Eq.CCEloss}
    \underset {\theta} { \operatorname {arg\,min} } \, \frac{1}{n}\cdot\frac{1}{\left\lvert s_i \right\rvert }\sum_{i = 1}^{n}\sum_{j \in s_i} -\log p_j(f(\boldsymbol{x}_i;\theta)).
\end{equation}
Following \cite{lv2021robustness}, if the dataset has $C$ classes, the $p_j(f(\boldsymbol{x}_i;\theta))$ can be specified as:
\begin{equation}
    p_j(f(\boldsymbol{x}_i;\theta)) = \frac{e^{f_{j}(\boldsymbol{x}_i;\theta)}}{\sum_{k=1}^C e^{f_{k}(\boldsymbol{x}_i;\theta)}},
\end{equation}
where $f(\cdot)_j$ is the output for $j$-th class. 
More specifically, $p_j(f(\cdot))$ is the $j$-th class probability of classifier $f(\cdot)$'s output.

Since the classifier will fit more unreliable labels after the early learning stage, the classifier $f(\cdot; \theta)$ should only be trained for several epochs $\beta$. 
Formally speaking, let $\tilde{D}_{R}^{i}$ be the reliable subset of $i$-th separation step. At the very beginning, set $\tilde{D}_{R}^{0}=\tilde{D}$.
Then let $\tilde{D}_{U}=\{\boldsymbol{x}_i\}_{i=1}^{m}$ denotes instances excluded from the reliable subset. 
$\theta_{j}^{i}$ is the parameters for $i$-th step separation's $j$-th training epoch.
Note that $\theta_{0}^{i}$ denotes the randomly initialized parameters in $i$-th separation step.

The RS algorithm can be described as follows: for the $i$-th step, the parameters $\theta_{0}^{i}$ are randomly initialized.
Then we train $f(\cdot;\theta)$ for several epochs $\beta$, get parameters $\theta_{\beta}^{i}$. 
After that, we retrieve the final epoch (i.e. $\beta$-th epoch) training losses for each sample and sorted them by loss value in descending order.
Simultaneously, instances that are correlated with the top-$\gamma$ ($0 < \gamma < 1$) maximum loss values are shifted to the $\tilde{D}_{U}^{i}$.
Following $\lambda$ steps of separation, we will acquire a reliable subset $\tilde{D}_{R}^{\lambda}$ and an unreliable subset $\tilde{D}_{U}^{\lambda}$.

If the unreliable rate $\mu$ on dataset is known, the $\lambda$ can be estimated by $\mu$ directly. 
However, it is generally not feasible in real-world settings.
Given this reality, a self-adaptive strategy has been devised to accommodate various levels of unreliability.
Intuitively, as the count of unreliable samples descending in $\tilde{D}_{R}^{i}$, the accuracy on validation or test set will increase.
But at the latter phase, most of the unreliable samples have been removed and samples in $\tilde{D}_{R}^{i}$ are totally reliable.
As the value of $\lvert \tilde{D}_{R}^{\lambda} \rvert$ goes down, the accuracy on the validation or test set diminishes.  
Given the need for addressing the limitations of unknown $\mu$, we propose a self-adaptive RS algorithm that incorporates an early-stopping technique to terminate the process of separation at an appropriate time. 
The self-adaptive strategy dictates that, should the accuracy on the validation set cease to improve over $\varphi$ consecutive epochs, the separation process will be terminated. This leads to the final determination of $\tilde{D}_{R}^{\lambda}$ and $\tilde{D}_{U}^{\lambda}$, with $\varphi$ representing the separation patience.
The details of self-adaptive RS algorithm is exhibited in Algorithm \ref{alg:RS}.

\subsection{Pilot Experiments}
\label{pilot_study}

\begin{figure}[t]
    \centering
    \includegraphics[width=0.95\columnwidth]{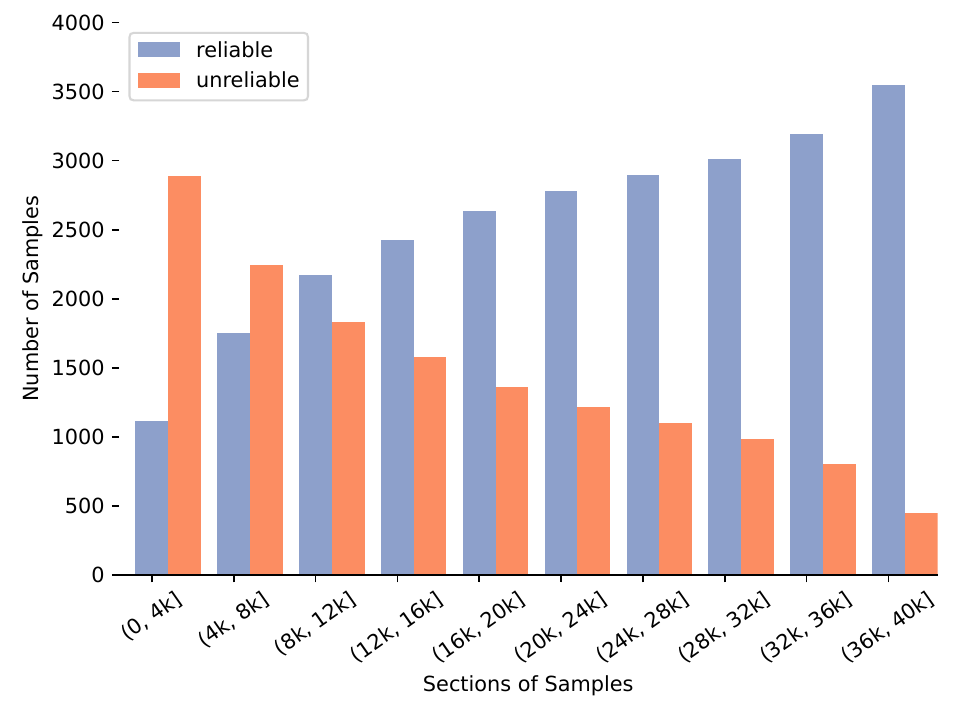}
    \caption{Number of samples for ten consecutive sections sorted by loss values in descending order. 
    In this figure, orange bars represent the number of unreliable samples and blue bars represent the number of reliable samples.
    For the first section $(0, 4\text{K}]$, which contains 4000 samples and almost 3000 samples are unreliable.
    But in the last section $(36\text{K}, 40\text{K}]$, it's the exact opposite of that.
    }
    \label{sample_num}
\end{figure}

\begin{figure}[t]
    \centering
    \includegraphics[width=0.95\columnwidth]{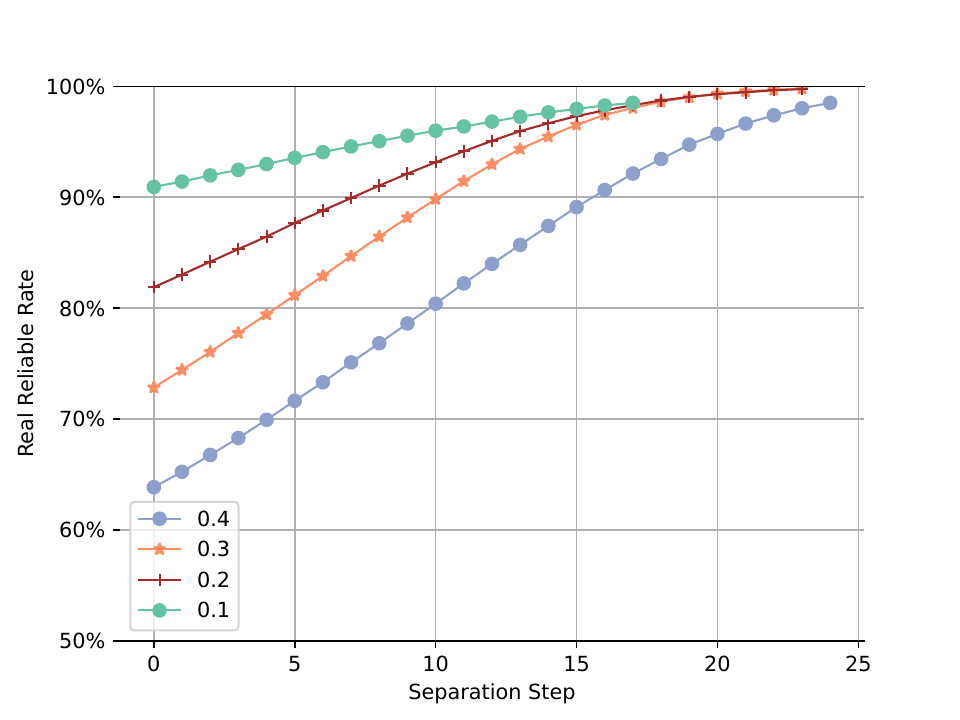}
    \caption{Real reliable rate variation on CIFAR-10 training set as separation step goes up in four different unreliable rate settings.
    The real reliable rate nearing 100\% means the subset $\tilde{D}_{R}^{\lambda}$ is almost reliable.
    }
    \label{select_acc}
\end{figure}

In order to validate the efficacy of the self-adaptive RS method put forth, a series of pilot experiments were conducted to assess the ability of the method to identify and exclude unreliable samples, resulting in the formation of a reliable subset.

To explore this idea, we generate UPLL dataset on CIFAR-10 \cite{cifar10} at first with four different unreliable rates 0.1, 0.2, 0.3 and 0.4.
Then annotate reliable or unreliable for each sample by the true label in or not in the candidate label set respectively.
The partial rate is fixed as 0.1 in the pilot experiments.
MultiLayer Perceptron (MLP) is used as backbone since complex networks will overfit unreliable samples faster than plain networks. 
The Categorical Cross Entropy (CCE) loss \cite{lv2021robustness} is utilized to train the classifier.
More generation process of the dataset see the section \ref{Experiments} below for a detailed description.

At first, we train the network for 5 epochs.
In the 5-th epoch, the samples in training set are sorted by loss value in descending order.
The experimental results are reported in Figure \ref{sample_num}.
The orange bars represent the number of unreliable samples and the blue bars represent the number of reliable samples.
The 40K samples are divided into 10 sections and each section contains 4K samples.
As is shown, the unreliable samples in the first section occupy a larger proportion than the reliable samples. 
But in the last section, it's exactly the opposite of that.

It is concluded that the section with higher loss value will contain more unreliable samples than the one with lower loss value.
Motivated by this finding, we try to exclude top-3\% samples every 5-epoch as a separation step with four different unreliable rates $\{0.1, 0.2, 0.3, 0.4\}$. 
Then record the variation of real reliable rate on the training set.
As shown in Figure \ref{select_acc}, the proportion of reliable samples increases as the number of separation steps increases.
That is to say, our self-adaptive RS method can effectively exclude unreliable samples and then get a highly reliable subset.
With relatively low unreliability, the model is able to achieve a higher level of accuracy. 
In the following, a framework is designed to learn from these two subsets.

\subsection{The Overall Framework}

The overall framework of UPLLRS consists of two stages.
Firstly, the self-adaptive RS component recursively separates $\tilde{D}$, obtaining a reliable subset $\tilde{D}_{R}^{\lambda}$ and an unreliable subset $\tilde{D}_{U}^{\lambda}$ as a result.
Then, the disambiguation strategy, in conjunction with a semi-supervised learning approach, induces the model from both $\tilde{D}_{R}^{\lambda}$ and $\tilde{D}_{U}^{\lambda}$.
Ultimately, we arrive at a well-trained classifier $g(\cdot;\Theta)$.

\begin{algorithm}[tb]
    \caption{UPLLRS Algorithm with General Solution}
    \label{alg:UPLLRS}
    \textbf{Input}: Network $g(\cdot; \omega)$ with parameters $\omega$.
    Unreliable partial label dataset $\tilde{D}$ and validation set $V$. 
    Max training epochs $T$.   \\
    \textbf{Output}: Parameters $\omega$ for $g(\cdot)$.

    \begin{algorithmic}[1] 
    \STATE Obtain reliable subset $\tilde{D}_{R}^{\lambda}$ and 
    unreliable partial dataset $\tilde{D}_{U}^{\lambda}$ by executing Algorithm \ref{alg:RS};
    \STATE Randomly initialize $\omega$.
    \FOR{$i \leftarrow 1$ to $T$}
        \STATE Train $g(\cdot;\omega)$ from $\tilde{D}_{R}^{\lambda}$;
        \STATE Calculate loss according Eq. \ref{Eq.proden};
        \STATE Update $\mathrm{w}_{i j}$ according Eq. \ref{Eq.weight};
        \STATE Use $g(\cdot;\omega)$ and get pseudo labels $R$ which over threshold $\tau$ on dataset $\tilde{D}_{U}^{\lambda}$;
        \STATE Add pseudo labels $R$ and corresponding instances to reliable dataset $\tilde{D}_{R}^{\lambda}$ and remove it from $\tilde{D}_{U}^{\lambda}$;
    \ENDFOR
    \STATE \textbf{return} $\omega$.

    \end{algorithmic}
\end{algorithm}

\subsubsection{General Solution}
After the recursive separation stage, the dataset is split into the reliable subset $\tilde{D}_{R}^{\lambda}$ and 
unreliable subset $\tilde{D}_{U}^{\lambda}$. 
As the unreliable rate $\mu$ goes up, $\tilde{D}_{R}$ will remain fewer and fewer samples while $\tilde{D}_{U}$ will collect more and more unreliable samples.
In order to fully leverage the information contained within the $\tilde{D}_{U}$, we employ pseudo-labeling technique for instances in $\tilde{D}_{U}^{\lambda}$. 
After the completion of each epoch, the model is evaluated on $\tilde{D}_{U}^{\lambda}$ and high-confidence samples are added to $\tilde{D}_{R}^{\lambda}$. 
More specifically, the pseudo label for each instance $\boldsymbol{x}_{i}$ is given by:
\begin{equation}
    u_i = \arg \max( p_{g}(\boldsymbol{x}_{i})),
\end{equation}
where $p_{g}$ denotes the model's predicted class distribution.
We only retain the pseudo labels which satisfy $max(p_{g}(\boldsymbol{x}_{i})) \geq \tau$,
where $\tau$ is a threshold. It is fixed as 0.95 in our experiments.

Although the $\tilde{D}_{R}^{\lambda}$ is reliable, the labels therein are ambiguous. 
Adopting the disambiguation method PRODEN \cite{lv2020progressive}, 
the weighted loss can be written as:
\begin{equation}
    \mathcal{L}=\frac{1}{m} \sum_{i=1}^{m} \sum_{j=1}^{C} \mathrm{w}_{i j} \mathcal{L}_{\text{CCE}}\left(g_{j}\left(\boldsymbol{x}_{i}\right), s_{i}\right),
    \label{Eq.proden}
\end{equation}
in which $\mathrm{w}_{i j}$ is the confidence of the $j$-th class being consistent with the concealed true class for the $i$-th instance.
It is estimated by the output of classifier $g(\cdot; \omega)$, which is defined as:
\begin{equation}
    \mathrm{w}_{i j}=\left\{\begin{array}{cc}
    g_{j}\left(\boldsymbol{x}_{i}\right) / \sum_{k \in s_{i}} g_{k}\left(\boldsymbol{x}_{i}\right) & \text { if } j \in s_{i}, \\
    0 & \text {otherwise},
    \end{array}\right.
    \label{Eq.weight}
\end{equation}
where $g_j(\cdot)$ is the $j$-th coordinate of $g(\cdot)$.
For initialization, the weights are uniform, i.e. if $j \in s_i$, 
$w_{i j}=1 / \lvert s_i \rvert $, otherwise $w_{i j}=0$.
The algorithm of the overall framework is illustrated in Algorithm \ref{alg:UPLLRS}.

\begin{table*}[htbp]
    \centering
    \resizebox{0.99\textwidth}{!}{
        \begin{tabular}{c|c|c|cccccccc}
            \midrule
            Dataset & \multicolumn{1}{c}{$\eta$} & $\mu$ & Ours  & RABS  & PiCO  & CR-DPLL & PRODEN & RC    & CC    & LWS \\
            \midrule
            \multirow{9}[6]{*}{CIFAR-10} & 0.1   & 0.1   & \textbf{95.16} $\pm$ 0.10\% & 83.87 $\pm$ 0.29\% & 91.35 $\pm$ 0.14\% & 93.49 $\pm$ 0.26\% & 79.77 $\pm$ 0.61\% & 79.96 $\pm$ 0.46\% & 78.91 $\pm$ 0.61\% & 85.83 $\pm$ 0.76\% \\
                & 0.1   & 0.3   & \textbf{94.65} $\pm$ 0.23\% & 77.75 $\pm$ 0.62\% & 87.66 $\pm$ 0.22\% & 90.65 $\pm$ 0.20\% & 67.80 $\pm$ 1.38\% & 69.46 $\pm$ 1.02\% & 67.52 $\pm$ 2.11\% & 19.95 $\pm$ 4.22\% \\
                & 0.1   & 0.5   & \textbf{93.12} $\pm$ 0.92\% & 65.09 $\pm$ 0.57\% & 82.47 $\pm$ 0.38\% & 85.65 $\pm$ 0.38\% & 51.07 $\pm$ 1.49\% & 54.75 $\pm$ 1.57\% & 52.37 $\pm$ 2.95\% & 16.65 $\pm$ 1.35\% \\
                \cmidrule{2-11}      & 0.3   & 0.1   & \textbf{94.32} $\pm$ 0.21\% & 53.13 $\pm$ 0.90\% & 90.50 $\pm$ 0.24\% & 92.92 $\pm$ 0.15\% & 77.12 $\pm$ 0.32\% & 75.39 $\pm$ 0.31\% & 75.37 $\pm$ 0.61\% & 83.92 $\pm$ 0.35\% \\
                & 0.3   & 0.3   & \textbf{93.85} $\pm$ 0.31\% & 41.61 $\pm$ 2.11\% & 86.37 $\pm$ 0.37\% & 88.80 $\pm$ 0.19\% & 62.06 $\pm$ 0.69\% & 61.87 $\pm$ 1.63\% & 62.91 $\pm$ 1.20\% & 78.33 $\pm$ 0.68\% \\
                & 0.3   & 0.5   & \textbf{91.16} $\pm$ 0.67\% & 30.33 $\pm$ 1.61\% & 79.87 $\pm$ 0.51\% & 82.06 $\pm$ 0.34\% & 44.38 $\pm$ 0.97\% & 47.13 $\pm$ 0.62\% & 45.75 $\pm$ 2.31\% & 24.16 $\pm$ 2.34\% \\
            \cmidrule{2-11}      & 0.5   & 0.1   & \textbf{92.47} $\pm$ 0.19\% & 31.62 $\pm$ 2.31\% & 89.48 $\pm$ 0.38\% & 91.88 $\pm$ 0.32\% & 73.30 $\pm$ 0.07\% & 68.17 $\pm$ 0.55\% & 71.03 $\pm$ 0.33\% & 70.46 $\pm$ 3.00\% \\
                & 0.5   & 0.3   & \textbf{91.55} $\pm$ 0.38\% & 27.88 $\pm$ 2.58\% & 84.48 $\pm$ 0.33\% & 86.78 $\pm$ 0.54\% & 57.25 $\pm$ 0.98\% & 54.55 $\pm$ 0.64\% & 54.69 $\pm$ 1.64\% & 58.31 $\pm$ 4.76\% \\
                & 0.5   & 0.5   & \textbf{89.56} $\pm$ 0.50\% & 24.48 $\pm$ 2.77\% & 74.68 $\pm$ 1.21\% & 78.31 $\pm$ 0.41\% & 42.99 $\pm$ 0.80\% & 42.43 $\pm$ 1.17\% & 36.96 $\pm$ 1.78\% & 40.23 $\pm$ 4.16\% \\
            \midrule
            \midrule
            \multirow{9}[6]{*}{CIFAR-100} & 0.01  & 0.1   & \textbf{75.73} $\pm$ 0.41\% & 27.38 $\pm$ 1.42\% & 67.94 $\pm$ 0.52\% & 74.22 $\pm$ 0.41\% & 55.68 $\pm$ 0.49\% & 56.08 $\pm$ 0.37\% & 55.35 $\pm$ 0.76\% & 5.37 $\pm$ 0.61\% \\
                & 0.01  & 0.3   & \textbf{71.72} $\pm$ 0.39\% & 17.56 $\pm$ 0.73\% & 62.12 $\pm$ 0.35\% & 68.56 $\pm$ 0.37\% & 45.31 $\pm$ 0.63\% & 44.80 $\pm$ 1.20\% & 44.95 $\pm$ 0.74\% & 3.61 $\pm$ 0.91\% \\
                & 0.01  & 0.5   & \textbf{66.40} $\pm$ 0.21\% & 11.73 $\pm$ 0.62\% & 54.84 $\pm$ 0.40\% & 61.93 $\pm$ 0.38\% & 32.87 $\pm$ 0.90\% & 32.55 $\pm$ 1.17\% & 33.62 $\pm$ 0.81\% & 3.17 $\pm$ 0.72\% \\
                \cmidrule{2-11}      & 0.05  & 0.1   & \textbf{74.73} $\pm$ 0.24\% & 31.65 $\pm$ 0.79\% & 66.67 $\pm$ 0.46\% & 73.34 $\pm$ 0.43\% & 52.05 $\pm$ 0.90\% & 50.04 $\pm$ 0.27\% & 52.02 $\pm$ 0.35\% & 16.11 $\pm$ 4.07\% \\
                & 0.05  & 0.3   & \textbf{70.31} $\pm$ 0.22\% & 21.45 $\pm$ 0.74\% & 59.01 $\pm$ 0.61\% & 66.79 $\pm$ 0.75\% & 37.81 $\pm$ 0.89\% & 25.06 $\pm$ 0.75\% & 40.24 $\pm$ 0.84\% & 8.49 $\pm$ 0.92\% \\
                & 0.05  & 0.5   & \textbf{64.78} $\pm$ 0.53\% & 15.08 $\pm$ 1.18\% & 46.81 $\pm$ 0.69\% & 59.09 $\pm$ 0.76\% & 20.84 $\pm$ 1.25\% & 19.93 $\pm$ 0.92\%  & 26.08 $\pm$ 0.66\% & 6.65 $\pm$ 0.66\% \\
            \cmidrule{2-11}      & 0.1   & 0.1   & \textbf{73.20} $\pm$ 0.50\% & 25.55 $\pm$ 1.55\% & 45.44 $\pm$ 1.68\% & 72.08 $\pm$ 0.52\% & 44.07 $\pm$ 0.47\% & 38.70 $\pm$ 1.52\% & 47.81 $\pm$ 0.90\% & 49.91 $\pm$ 0.97\% \\
                & 0.1   & 0.3   & \textbf{68.60} $\pm$ 0.25\% & 16.99 $\pm$ 2.06\% & 35.89 $\pm$ 1.48\% & 64.70 $\pm$ 0.45\% & 25.66 $\pm$ 0.58\% & 21.26 $\pm$ 0.63\% & 34.02 $\pm$ 0.76\% & 18.11 $\pm$ 2.83\% \\
                & 0.1   & 0.5   & \textbf{60.66} $\pm$ 0.75\% & 10.80 $\pm$ 0.62\% & 22.57 $\pm$ 1.07\% & 52.34 $\pm$ 0.62\% & 13.61 $\pm$ 0.63\% & 12.89 $\pm$ 0.62\% & 20.63 $\pm$ 0.63\% & 9.52 $\pm$ 0.46\% \\
            \midrule
            \midrule
            \end{tabular}%
        }
    \caption{Test accuracy (mean$\pm$std) on CIFAR-10 and CIFAR-100 synthesized dataset. The best results are highlighted in bold.}
    \label{tab:cifar10cifar100}
\end{table*}

\subsubsection{Augmented Solution for Image Datasets}
The general solution is able to handle both image and non-image datasets.
As for image datasets, augmentation is an important procedure in classification tasks \cite{Shorten2019augmentation}.
Both $\tilde{D}_{R}^{\lambda}$ and $\tilde{D}_{U}^{\lambda}$ can employ image augmentation strategies to further enhance their performance.
For the $\tilde{D}_{R}^{\lambda}$, the consistency regularization based method CR-DPLL \cite{wu2022revisiting} is capable of being employed which takes advantage of image augmentation and achieves promising performance on PLL. 
To be specific, the $\mathcal{L}_{\text{PLL}}$ can be written as:
\begin{equation}
    \mathcal{L}_{\text{PLL}}=\mathcal{L}_{\text {Sup }}(\boldsymbol{x}, s)+ \pi(t) \boldsymbol{\Psi}(\boldsymbol{x}, s),
    \label{Eq.RCRPLL}
\end{equation}
where $\mathcal{L}_{\text {Sup }}(\boldsymbol{x}, s)=-\sum_{k \notin s} \log \left(1-g_k(\boldsymbol{x})\right)$ and $\boldsymbol{\Psi}(\boldsymbol{x}, s)=\sum_{\boldsymbol{z} \in \mathcal{A}(\boldsymbol{x})} \mathrm{KL}(s \| g(\boldsymbol{z}))$. 
$\mathrm{KL}(\cdot)$ denotes the Kullback-Leibler divergence and $\mathcal{A}(\boldsymbol{x})$ denotes the set of random augmented versions of instance $\boldsymbol{x}$.
$\pi(t) = \min\{t \pi/T', \pi\}$ is a dynamic balancing factor, where $t$ is the current epoch and $T'$ is a constant.  More specifically, the factor is increased to $\pi$ at the $T'$-th epoch, and thereafter maintained at a constant value of $\pi$ until the end of the training.
Meanwhile, the label weights are iteratively updated every epoch, more details are presented in Appendix A.1.

As for $\tilde{D}_{U}^{\lambda}$, a semi-supervised learning method \cite{sohn2020fixmatch} can be leveraged to extract potential valuable information in the unreliable instances.
Specifically, for the images in the unreliable subset $\tilde{D}_{U}^{\lambda}$, the pseudo labels generated by the model's prediction where the images is weakly augmented.
Next, we selectively preserve the samples whose pseudo labels satisfy the condition of $\max(p_{g}(x_{i}^{w})) \geq \tau$. 
Then the model is trained to predict the pseudo labels when fed with a strongly-augmented version of the same image.
Thus, the loss function for unreliable subset $\tilde{D}_{U}^{\lambda}$ takes the following form:
\begin{equation}
    \resizebox{.98\linewidth}{!}{$
    \mathcal{L}_{\text{U}}=\frac{1}{n-m} \sum_{i=1}^{n-m} \mathds{1}(\max(p_{g}(x_{i}^{w})) \geq \tau) \mathcal{L}_{\text{CE}}\left(g\left(x_{i}^{w}\right), g\left(x_{i}^{s}\right)\right),
    $}
    \label{Eq.aug}
\end{equation}
where $x_{i}^{w}$ and $x_{i}^{s}$ are the weak and strong augmentation of $x_{i}$ respectively.
$\mathds{1}(\cdot)$ is an indicator function and $\mathcal{L}_{\text{CE}}$ represents corss-entropy loss.
Consequently, the objective of UPLLRS is as follows:
\begin{equation}
    \mathcal{L}=\mathcal{L}_{\text{PLL}}+ \xi \mathcal{L}_{\text{U}},
    \label{Eq.loss}
\end{equation}
where $\xi$ is a scalar hyperparameter.

\section{Experiments}
\label{Experiments}
\label{experiments}

\begin{table*}[htbp]
    \centering
    \resizebox{0.9\textwidth}{!}{
        \begin{tabular}{c|cccc|cccc}
            \midrule
                & \multicolumn{4}{c|}{Dermatology} & \multicolumn{4}{c}{20Newsgroups} \\
            \midrule
            $\eta$ & 0.1   & 0.1   & 0.3   & 0.3   & 0.1   & 0.1   & 0.3   & 0.3 \\
            $\mu$ & 0.3   & 0.5   & 0.3   & 0.5   & 0.3   & 0.5   & 0.3   & 0.5 \\
            \midrule
            Ours & \textbf{96.06} $\pm$ 1.31\% & \textbf{89.75} $\pm$ 1.31\% & \textbf{91.80} $\pm$ 2.74\% & \textbf{87.87} $\pm$ 6.52\% & \textbf{72.27} $\pm$ 1.55\% & \textbf{65.41} $\pm$ 0.96\% & \textbf{61.47} $\pm$ 1.13\% & \textbf{47.98} $\pm$ 1.51\% \\
            RABS  & 78.36 $\pm$ 6.33\% & 47.86 $\pm$ 6.67\% & 58.69 $\pm$ 5.42\% & 46.88 $\pm$ 8.95\% & 64.18 $\pm$ 1.00\% & 50.99 $\pm$ 0.79\% & 31.97 $\pm$ 1.09\% & 23.45 $\pm$ 0.53\% \\
            PRODEN & 82.95 $\pm$ 4.34\% & 62.29 $\pm$ 10.52\% & 77.70 $\pm$ 3.96\% & 60.98 $\pm$ 8.88\% & 64.08 $\pm$ 0.43\% & 50.69 $\pm$ 1.36\% & 58.79 $\pm$ 1.03\% & 42.96 $\pm$ 0.90\% \\
            RC    & 79.34 $\pm$ 4.82\% & 61.63 $\pm$ 7.30\% & 76.39 $\pm$ 7.93\% & 54.75 $\pm$ 7.86\% & 63.23 $\pm$ 0.70\% & 48.33 $\pm$ 0.84\% & 56.09 $\pm$ 0.71\% & 39.43 $\pm$ 1.06\% \\
            CC    & 83.93 $\pm$ 3.34\% & 60.65 $\pm$ 9.72\% & 81.97 $\pm$ 5.18\% & 55.41 $\pm$ 8.52\% & 62.39 $\pm$ 1.05\% & 48.10 $\pm$ 0.39\% & 54.55 $\pm$ 0.88\% & 37.19 $\pm$ 1.38\% \\
            LWS   & 83.28 $\pm$ 4.90\% & 74.42 $\pm$ 12.33\% & 77.05 $\pm$ 3.28\% & 63.93 $\pm$ 7.33\% & 40.17 $\pm$ 4.64\% & 24.99 $\pm$ 2.16\% & 11.20 $\pm$ 1.08\% & 9.14 $\pm$ 0.64\% \\
            \midrule
            \midrule
            \end{tabular}
    }
    \caption{Test accuracy (mean$\pm$std) on UCI synthesized dataset.}
    \label{tab:uci}
\end{table*}

A comprehensive set of experiments were conducted to evaluate the performance of our method under varying levels of partial and unreliable labeling. The results demonstrate that our approach achieves state-of-the-art accuracy on tasks involving UPLL.

\subsection{Datasets and Implementation Details}
\subsubsection{Datasets}
We utilize two commonly employed image datasets, CIFAR-10 and CIFAR-100 \cite{cifar10}, 
as the basis for synthesizing our UPLL dataset.
Besides, we also utilize two additional datasets Dermatology and 20Newsgroups from UCI machine learning Repository \cite{Dua:2019} to further validate the effectiveness of our proposed method.
In our experiments, the datasets are partitioned into training, validation, test set in a 4:1:1 ratio. 
Further elaboration can be found in Appendix A.2.

Following the confusing strategy in \cite{lv2021robustness}, the ground-truth labels in the raw dataset are corrupted initially and then generate partial labels by the flipping process.
That is to say, for an instance $x$ with the ground-truth label $y=i, i \in \mathcal{Y}$, it has a fixed probability $1-\mu$ do not make any operation.
But it has a probability $\kappa$ to flip into $j$, where $j \in \mathcal{Y}, j \neq i, \kappa=\mu / (C-1)$.
The unreliable label is called $\tilde{y}_i $. 
Subsequently, $\tilde{y}_i$ is considered as the true label for generating the candidate label set, employing a uniform partial labeling with probability $\eta$ in accordance with the approach presented in \cite{lv2020progressive}, where $\eta$ denotes the partial rate.

\subsection{Baselines}
In order to demonstrate the efficacy of our proposed method and to gain insight into its underlying characteristics, we conduct comparisons with seven benchmark methods including one UPLL method and six state-of-the-art PLL methods: 1) RABS \cite{lv2021robustness}: An unreliable PLL method that proved the robustness of Average-Based Strategy (ABS) with bounded loss function in mitigating the impact of unreliability. In our experiment, the Mean Average Error (MAE) loss is chosen as the baseline.
2) PiCO \cite{wang2022pico}: A PLL method combines the idea of contrastive learning and class prototype-based label disambiguation method.
3) CR-DPLL \cite{wu2022revisiting}: A deep PLL method based on consistency regularization.
4) PRODEN \cite{lv2020progressive}: A PLL method which progressively identifies true labels in candidate label sets. 
5) RC \cite{feng2020provably}: A risk-consistent method for PLL which employs importance re-weighting strategy.
6) CC \cite{feng2020provably}: A classifier-consistent method for PLL using transition matrix to form an empirical risk estimator.
7) LWS \cite{wen2021leveraged}: A PLL method utilizing Leveraged weighted (LW) loss which balances the trade-off between losses on partial labels and others.

Note that two partial label learning methods PiCO \cite{wang2022pico} and CR-DPLL \cite{wu2022revisiting} are not suitable on the Dermatology and 20Newsgroups. 
More details can be found in Appendix A.2.

\begin{table*}[t]
    \centering
    \resizebox{1.5\columnwidth}{!}{
        \begin{tabular}{c|c|c|c|c|c|c|c|c}
            \toprule
                & \multicolumn{1}{c}{} &       & \multicolumn{3}{c|}{CIFAR-10 ($\eta=0.3$)} & \multicolumn{3}{c}{CIFAR-100 ($\eta=0.05$)} \\
            \midrule
            Ablation & RS    & $\tilde{D}_{U}^{\lambda}$ & $\mu=0.1$ & $\mu=0.3$  & $\mu=0.5$  &  $\mu=0.1$ & $\mu=0.3$  & $\mu=0.5$  \\
            \midrule
            UPLLRS & \Checkmark & \Checkmark & \textbf{94.32} $\pm$ 0.21\% & \textbf{93.85} $\pm$ 0.31\% & \textbf{91.16} $\pm$ 0.67\% & \textbf{74.73} $\pm$ 0.24\% & 70.31 $\pm$ 0.22\% & \textbf{64.78} $\pm$ 0.53\% \\
            UPLLRS w/o $\tilde{D}_{U}^{\lambda}$   & \Checkmark & \XSolid & 93.07 $\pm$ 0.08\% & 92.48 $\pm$ 0.27\% & 89.81 $\pm$ 0.39\% & 74.35 $\pm$ 0.52\% & \textbf{70.38} $\pm$ 0.50\% & 64.56 $\pm$ 0.40\% \\
            UPLLRS w/o RS & \XSolid & \XSolid & 92.92 $\pm$ 0.15\% & 88.80  $\pm$ 0.19\%  & 82.06 $\pm$ 0.34\% & 73.34 $\pm$ 0.43\% & 66.79 $\pm$ 0.75\% & 59.09 $\pm$ 0.76\% \\
            \bottomrule
            \bottomrule
            \end{tabular}%
    }
    \caption{The impact of RS and Unreliable Subset $\tilde{D}_{U}^\lambda$ on accuracy (mean$\pm$std).}
    \label{tab:ablation_tab}
\end{table*}

\subsubsection{Implementation Details}
For the first stage (i.e. self-adaptive RS), a 5-layer perceptron (MLP) is utilized to separate samples with CCE \cite{lv2021robustness} loss.
The learning rate is 0.1, 0.18, 0.1; small epochs $\beta=5, 6, 5$; separation rate $\gamma=0.03, 0.005, 0.03$; on the CIFAR-10, CIFAR-100 and UCI datasets respectively. 
Max separation step $\lambda=\lfloor\log_{1-\gamma}{0.3}\rfloor$.
As for the second stage, we employ different backbones for different datasets.
On CIFAR-10 dataset and CIFAR-100 dataset, we use WideResNet$28\times 2$ \cite{Zagoruyko2016WRN} as the predictive model, and we employ the Augmented Solution.
On the UCI datasets, a 5-layer perceptron (MLP) is employed, and we utilize the General Solution.
The learning rate is $5e-2$ and the weight decay is $1e-3$;
$\xi$ is set as $2$ on CIFAR-10 and $0.3$ on CIFAR-100.
We implement the data augmentation technique following the "strong augmentation" in CR-DPLL \cite{wu2022revisiting}.
This processing is applied on Ours, PRODEN, RC, CC, LWS.
As for PiCO and CR-DPLL, the augmentation setups followed recommended setting in the previous works.

The optimizer in our experiment is Stochastic Gradient Descent (SGD) \cite{robbins1951stochastic} in which momentum is set as $0.9$.
For the learning rate scheduler, we use a cosine learning rate decay \cite{loshchilov2016sgdr}.
Otherwise, each model is trained with maximum epochs $T=500$ and employs early stopping strategy with patience 25.
In other words, if the accuracy does not rise in validation set $V$ for 25 epochs, the training process will be stopped.
All experiments are conducted on NVIDIA RTX 3090.
What's more, the implementation of our method is based on PyTorch \cite{paszke2019pytorch} framework.
We report final performance using the test accuracy corresponding to the best accuracy on validation set for each run.
Finally, we report the mean and standard deviation based on five independent runs with different random seeds.

\subsection{Experiment Results}
Table \ref{tab:cifar10cifar100} reports the experimental results on CIFAR-10 and CIFAR-100 synthesized datasets.
As is shown, our UPLLRS method outperforms all compared methods.
The improvements are particularly pronounced in scenarios with high levels of unreliability.
Take $\eta=\{0.1, 0.3, 0.5\}, \mu=0.5$ on CIFAR-10 dataset as an instance, our method improves by 7.47\%, 9.1\%, 11.25\% respectively compared with the second-best methods.
It is worth noting that our method exhibits a minimal decline in accuracy as the unreliable rate $\mu$ increases.
For example, for $\eta=0.5$, the accuracy for $\mu=0.1$ is 92.47\% and $\mu=0.5$ is 89.56\%, only 2.91\% accuracy drop.
In contrast, the second-best method's accuracy drop up to 13.57\%.
UPLLRS also achieves the best performance and significantly outperforms other compared methods on the CIFAR-100 synthesized dataset.
For the settings with $\eta=0.1, \mu=0.5$, our UPLLRS also achieves 60.66\% outperforming second-best 8.32\%. 
Conversely, other methods either exhibit poor performance or fail to converge.

We further evaluate the performance of UPLLRS on non-image datasets Dermatology and 20Newsgroups. 
Table \ref{tab:uci} reports the experimental results on it.
Our method demonstrates a clear advantage and surpasses all the methods evaluated in comparison.
For instance, in the condition of $\eta=0.3, \mu=\{0.3, 0.5\}$ on Dermatology, our method exhibited a 9.83\%, 23.94\% over the second-best method respectively.
Furthermore, during experimentation with $\eta=0.1$, our method exhibited a drop of 11.29\% when varying the $\mu$ from $0.3$ to $0.5$.
Hence, our method has noticeable resistance to unreliability. 
In contrast, other methods exhibit a significant decline.

\subsection{Ablation Study}

In this subsection, we present the results of our ablation study which serve to demonstrate the efficacy of the components of our UPLLRS method: RS and Unreliable Subset $\tilde{D}_{U}^\lambda$.
The experiments are conducted on CIFAR-10 dataset with $\eta=0.3, \mu=\{0.1, 0.3, 0.5\}$ and CIFAR-100 dataset with $\eta=0.05, \mu=\{0.1, 0.3, 0.5\}$. 
Other hyperparameter settings are consistent with those utilized in the primary experiments.
Besides, analysis on hyperparameter $\xi$ and $\gamma$ are elaborated in detail in the Appendix A.3.

Here, we conduct ablation studies on the individual components to investigate their contributions.
Two variants are selected: 1) Without RS. 
That implies that the corrupted dataset will not be partitioned into subsets, but rather utilized directly to induce the final classifier.
2) Without unreliable subset $\tilde{D}_{U}^{\lambda}$. 
It can be stated that the final classifier is directly trained on the reliable subset.
All other parameters are held constant as in the primary experiment.
As shown in Table \ref{tab:ablation_tab}, It is apparent that the contribution of RS surpasses that of $\tilde{D}_{U}^{\lambda}$.
Take CIFAR-10 with $\mu=0.5$ as an instance, it was observed that the variant without the utilization of the $\tilde{D}_{U}^\lambda$ experienced a mere 1.35\% decline in performance compared to the full UPLLRS model. 
However, when comparing the variant without the self-adaptive RS to the variant without the $\tilde{D}_{U}^\lambda$, a significant decline of 7.75\% was observed.
Furthermore, while utilizing the $\tilde{D}_{U}^\lambda$ on the CIFAR-100 dataset resulted in a slight increase, this can likely be attributed to the lower accuracy of pseudo-label generation from $\tilde{D}_{U}^\lambda$, as CIFAR-100 has a substantially larger number of classes than CIFAR-10.

\section{Conclusion}
In this work, we propose a novel two-stage framework named Unreliable Partial Label Learning with Recursive Separation (UPLLRS).
First, the self-adaptive recursive separation strategy 
is proposed to separate the training set into a reliable subset and an unreliable subset.
Second, a disambiguation strategy progressively identifies ground-truth labels in the reliable subset.
Meanwhile, the semi-supervised learning techniques are employed for the unreliable subset. 
Experimental results demonstrate that our method attains state-of-the-art performance, particularly exhibiting robustness in scenarios with high levels of unreliability.

\section*{Acknowledgments}
This work is supported by National Key R\&D Program of China (2018AAA0100104), the National Science Foundation of China (62206050, 62125602, and 62076063), China Postdoctoral Science Foundation (2021M700023), Jiangsu Province Science Foundation for Youths (BK20210220), Young Elite Scientists Sponsorship Program of Jiangsu Association for Science and Technology (TJ-2022-078), and the Big Data Computing Center of Southeast University.

\bibliographystyle{named}
\bibliography{shiyu}

\clearpage


\end{document}